\documentclass[10pt]{article}
\usepackage{ijcai22}

\usepackage{times}
\usepackage{soul}
\usepackage{url}
\usepackage[hidelinks]{hyperref}
\usepackage[utf8]{inputenc}
\usepackage[small]{caption}
\usepackage{graphicx}
\usepackage{amsmath}
\usepackage{amsthm}
\usepackage{booktabs}
\usepackage{algorithm}
\usepackage{algorithmic}
\usepackage{multirow}
\usepackage{wasysym}
\usepackage{multirow}
\usepackage{pifont}  
\usepackage[table,xcdraw]{xcolor}
\usepackage[capitalize,noabbrev]{cleveref}
\usepackage{array, makecell}
\usepackage{tablefootnote}
\newcommand{\cmark}{\ding{51}}  
\newcommand{\xmark}{\ding{55}}  

\definecolor{darkblue}{rgb}{0,0.08,0.45}
\hypersetup{
    colorlinks=true,
    linkcolor=darkblue,
    citecolor=darkblue,
    filecolor=darkblue,
    urlcolor=darkblue,
}

\hypersetup{
    pdfauthor={Julia Gusak,
               Daria Cherniuk,
               Alena Shilova,
               Alexander Katrutsa,
               Daniel Bershatsky,
               Xunyi Zhao,
               Lionel Eyraud-Dubois,
               Oleg Shlyazhko,
               Denis Dimitrov,
               Ivan Oseledets,
               Olivier Beaumont},
    pdftitle={Survey on Large Scale Neural Network Training},
    pdfsubject={Deep Learning},
    pdfkeywords={
        Machine Learning, Deep Learning, Rematertialization, Checkpointing, Offloading, Optimizers, Cross-Device Training
    },
}

\title{Survey on Large Scale Neural Network Training}

\author{
Julia~Gusak\textsuperscript{1,*}\and
Daria~Cherniuk\textsuperscript{1,*}\and
Alena~Shilova\textsuperscript{2,*}\and \\
Alexander~Katrutsa\textsuperscript{1}\and
Daniel~\mbox{Bershatsky}\textsuperscript{1}\and
Xunyi~Zhao\textsuperscript{3}\and
Lionel~\mbox{Eyraud-Dubois}\textsuperscript{3}\and
Oleg~Shlyazhko\textsuperscript{4}\and \\
Denis~Dimitrov\textsuperscript{4}\and
Ivan~Oseledets\textsuperscript{1}\and
Olivier~Beaumont\textsuperscript{3}
\affiliations
\textsuperscript{1}Skolkovo Institute of Science and Technology, Russia\\
\textsuperscript{2}Inria, University of Lille - CRIStAL, France \\
\textsuperscript{3}Inria, University of Bordeaux, France\\
\textsuperscript{4}Sber AI, Russia\\
\textsuperscript{*}equal contribution 
\emails
\{y.gusak, daria.cherniuk, a.katrutsa, d.bershatsky2, i.oseledets\}@skoltech.ru, \\
\{alena.shilova, xunyi.zhao, lionel.eyraud-dubois, olivier.beaumont\}@inria.fr, \\
\{omshlyazhko, dimitrov.d.v\}@sberbank.ru
}

\begin{document}

\maketitle

\begin{abstract}
Modern Deep Neural Networks (DNNs) require significant memory to store weight, activations, and other intermediate tensors during training. Hence, many models don’t fit one GPU device or can be trained using only a small per-GPU batch size. This survey provides a systematic overview of the approaches that enable more efficient DNNs training. We analyze techniques that save memory and make good use of computation and communication resources on architectures with a single or several GPUs. We summarize the main categories of strategies and compare strategies within and across categories. Along with approaches proposed in the literature, we discuss available implementations.
\end{abstract}

\section{Introduction}

Modern trends in the development of Deep Learning (DL) and Artificial Intelligence (AI) technologies involve the use of Deep Neural Networks (DNNs) to solve various problems of image, video, audio, natural language processing, content generation in the form of images or text in a given style and subject, etc.

The question we address in this survey is the following: given your model (which you do not want to rewrite) and your computation platform (which you do not want to change), what are the generic approaches that can allow you to perform the training efficiently? For training to be efficient, it must be feasible (the data must fit in memory), it must exploit the computational power of the resources well (the arithmetic intensity of the operations must be sufficient) and, in the parallel case, it must not be limited by too large data exchanges between the nodes. 
The efficiency of the training depends fundamentally on the efficient implementation of computational kernels on the computational resources (CPU, TPU, GPU) and on the efficient implementation of communications between GPUs and between different memories. In both cases, a lot of work has been done on the optimization of the arithmetic intensity of the computational kernels and on the efficient realization of collective communication operations over the hardware network. For the user, powerful profiling tools have been developed to identify hardware bottlenecks and can be used to decide which strategies described in this survey can be used to solve the problems of arithmetic intensity , memory and by controlling the volume of data exchanged.

The present survey covers generic techniques to cope with these limitations. If the computation cannot be performed a priori because the model, the optimizer states, and the activations do not fit in memory, there are techniques to trade memory for computation (re-materialization) or for data movements (activation and weight offloading), and it is also possible to compress the memory footprint by approximating optimizer states and gradients (compression, pruning, quantization). The use of parallelism (data parallelism, model parallelism, pipelined model parallelism) can also make it possible to distribute the memory requirements over several resources.
If the arithmetic intensity of the computations is not sufficient to fully exploit the GPUs and TPUs, it is generally because the size of the mini-batch is too small, and then the above techniques can also enable to increase the size of the mini-batch.
Finally, if the communications, typically induced by the use of data parallelism, are too expensive and slow down the computation, then other forms of parallelism can be used (model parallelism, pipelined model parallelism) and the compression of the gradients can allow to limit the volumes of exchanged data.

In this survey, we explain how these different techniques work, we describe the literature to evaluate and compare the proposed approaches and we also analyze the frameworks that allow to implement these techniques (almost) transparently. The different techniques that we consider, and their influence on communications, memory and computing efficiency are depicted in~\cref{tab:methods}.

\begin{table*}[t!]
    \centering
    \resizebox{\linewidth}{!}{
    \begin{tabular}{lrcccr}
        \toprule
        Method  & \makecell{$\#$ of\\ GPUs} & \makecell{Approx.\\ computations} & \makecell{Communication costs per iteration\\ activation values / weight values / \\ activation grads / weight grads} & \makecell{Batch size per \\ GPU  increase?} & \makecell{$\#$ of FLOP \\ per iteration}\\
        \midrule
        No data parallelism & 1 & baseline & baseline  & baseline  &  baseline \\
        Rematerialization & $\ge 1$ &  \xmark &  = / =\quad = / =  &  \cmark & $\uparrow$ \\
        Offloading: & & & & & \\
        \quad activations & $\ge 1$ &  \xmark &   $\uparrow$ / =\quad = / =& \cmark & = \\
        \quad weights & $\ge 1$&  \xmark & = / $\uparrow$ \quad = / $\uparrow$ \text{or} =  & \cmark & = \\
        \quad tensors in GPU cache & $\ge 1$&  \xmark & $\uparrow$ \text{or} = / $\uparrow$ \text{or} = \quad  = /  $\uparrow$ \text{or} = & \cmark & = \\
        \hline
        Approx. gradients: & &  & & & \\
        \quad lower-bit activation grad.\textsuperscript{**} & $\ge 1$ & \cmark &  $\downarrow$ / = \quad   = / = & \cmark & $\downarrow$ \text{or} =\\
        \quad\makecell[l]{approx. matmul\textsuperscript{**}} & $\ge 1$ & \cmark &  $\downarrow$ / = \quad  = / = & \cmark & $\updownarrow$\\
        \quad lower-bit weight grad\textsuperscript{**}  & $\ge 1$ & \cmark &   = / = \quad  = / $\downarrow$ & \cmark & $\downarrow$ \text{or} = \\
        \hline
        \hline
        Data parallelism\textsuperscript{*} &  $> 1$ &  \xmark  & baseline  & \xmark & = \\
        \quad Partitioning:  &   &  & & & \\
        \quad\quad optim. state  &  $> 1$ &  \xmark &  = / $\uparrow$  \quad   = / =  & \cmark & = \\
        \quad\quad + gradients  &  $> 1$ &  \xmark  &   = / $\uparrow$  \quad  = / = & \cmark & = \\
        \quad\quad + parameters &  $> 1$ &  \xmark & = / $\uparrow$ \quad  = / =  & \cmark & = \\
        \hline
        Model parallelism\textsuperscript{*}&  $> 1$ &  \xmark &   $\uparrow$ / = \quad $\uparrow$ / $\downarrow$ & \cmark & = \\
        Pipeline parallelism &  $> 1$ &  \cmark / \xmark & $\uparrow$ / = \quad $\uparrow$ / $\updownarrow$  & \cmark & = \\
        \bottomrule
        \multicolumn{6}{l}{\makecell[l]{$^*$\small We assume updates performed in synchronous way. If updates are asynchronous than for both data and model parallelism less gradients\\\small contribute to epoch update and number of epochs till convergence might be increased.}} \\
        \multicolumn{6}{l}{\makecell[l]{$^{**}$\small Communication channel in this case is a bus between a processing unit and a memory bank.}} \\
    \end{tabular}
    }
    \caption{Methods to train large neural networks. $\uparrow$ and $\downarrow$ correspond to the increase and decrease comparing to the baseline above. FLOP is floating-point operations.}
    \label{tab:methods}
\end{table*}

According to our taxonomy, we distinguish the following methods based on their purpose: reducing the memory footprint on a GPU is discussed in \cref{sec:one_gpu}, the use of parallel training for models that do not fit on a GPU is considered in  \cref{sec:parallelisms}, and the design of optimizers developed to train models stored on multiple devices is addressed in \cref{sec:optimizers}.

\section{Memory Usage Reduction on a Single GPU}\label{sec:one_gpu}

During the forward pass, neural networks store the activations necessary to perform backpropagation. In some cases, these activations can consume a substantial amount of memory, making training infeasible. There are two main approaches to reducing that memory footprint: rematerialization (also called checkpointing) and offloading.
Let us introduce the following notations: $F_i$/$B_i$ is a computation of forward/backward pass on module $i$; $F_i^n$ is an operation that computes the output of $F_i$ and forgets its input; $F_i^c$ computes the output of $F_i$ and keeps the input; $F_i^e$ computes the output of $F_i$, keeps the input, and all intermediate activations needed to later compute $B_i$. These different operations can easily be implemented in frameworks like PyTorch or Tensorflow.


\subsection{Rematerialization of Activations}

{\it Rematerialization} is a strategy that only stores a fraction of the activations during the forward pass and recomputes the rest during the backward pass. 
Rematerialization methods can be distinguished by what computational graphs they are dealing with.
The first group comes from Automatic Differentiation (AD), they find optimal schedules for homogeneous sequential networks (DNNs whose layers are executed sequentially and have the same computational and memory costs). The second group concentrates on transition model such as heterogeneous sequential networks (DNNs may be any sequential neural networks consisting of arbitrarily complex modules, e.g. CNNs, ResNet, some transformers), which adjust solutions from AD to heterogeneous settings. 
The final group concentrates on general graphs, but this problem becomes NP-complete in the strong sense and thus can be solved optimally only with ILP, otherwise approximately with various heuristics.

For homogeneous sequential networks, the binomial approach was proven to be optimal in \cite{Grimm1996} and implemented in REVOLVE~\cite{Griewank2000}. Compiler-level techniques have been proposed in~\cite{Siskind2018} to make it applicable to fully arbitrary programs. This results in a divide-and-conquer strategy, which however assumes that computations can be interrupted at arbitrary points, making it unsuitable for GPU computations. 
In the case of heterogeneous computation times and homogeneous memory costs the optimal strategy can be found with dynamic programming \cite{Griewank2008}. 
A direct adaptation of results for homogeneous chains was applied to RNNs in \cite{Gruslys2016}.

\cite{herrmann2019optimal} considered the most general case of heterogeneous sequential DNNs. 
This paper proposed a new modeling of the problem directly inspired by the data dependencies induced by the PyTorch framework. A dynamic programming approach is used that finds, for linear or linearized chains, the optimal strategy.

 Some methods can perform rematerialization for general graphs, though the exact approaches are exponentially expensive 
 (see Table~\ref{tab:remat}). For example, \cite{Jain2019} proposed an Integer Linear Program (ILP) to find the optimal rematerialization strategy suitable for an arbitrary Directed Acyclic Graph (DAG) structure. 
\cite{kirisame2020dynamic} presented a cheap dynamic heuristic DTR that 
relies on scores encouraging to discard (i) heavy tensors (ii) with a long lifetime and (iii) that can be easily recomputed. 


\begin{table*}[t!]
    \centering
    \resizebox{\linewidth}{!}{
    \begin{tabular}{lp{0.19\textwidth}p{0.115\textwidth}p{0.12\textwidth}p{0.11\textwidth}p{0.13\textwidth}}
        \toprule
        Paper  & Approach & Scope & Guarantees & Complexity & Implementation\\
        \midrule
        \cite{Griewank2000} & dynprog & \multirow{2}{*}{hom. seq.} & \multirow{2}{*}{optimal} & \multirow{2}{*}{$O(L^2 M)$} & \multirow{2}{*}{REVOLVE} \\
        \cite{Grimm1996} & closed-form &&&& \\
        \cite{Griewank2008} & dynprog & het. time, \newline hom. memory & optimal & $O(L^3 M)$ & -\\
        \cite{Siskind2018} & divide $\&$ conquer & compiler & -  & - & checkpointVLAD
        \\
        \hline
        \cite{Chen2016} & periodic & het. seq & heuristic & - & PyTorch \\ 
        \cite{Gruslys2016} & dynprog & RNN & - & $O(L^2M)$ & BPTT \\
        \cite{herrmann2019optimal} & dynprog & het. seq & optimal w/ \newline assumptions & $O(L^3 M)$ & Rotor\\
        \hline
        \multirow{2}{*}{\cite{Jain2019}} & ILP & \multirow{2}{*}{any} & optimal & NP-hard & \multirow{2}{*}{CheckMate}\\
         & rational LP & & heuristic & $O(EL)$ vars \& constraints & \\
        \multirow{2}{*}{\cite{kusumoto2019graph}} & exact dynprog & \multirow{2}{*}{any} & - & $O(T 2^{2L}))$ & \multirow{2}{*}{Recompute} \\
        & approx. dynprog & & heuristic & $O(T L^2)$ & \\
        \cite{kumar2019efficient} & tree-width decomposition & any & bounded & $2^{O(w)}L + O(wL\log L)$ & -\\
        \cite{kirisame2020dynamic} & greedy (priority scores) & any & heuristic & - & DTR \\
        \bottomrule
    \end{tabular}
    }
    \caption{Comparison of rematerialization strategies. The complexity is expressed with respect to number of modules $L$,  memory on GPU $M$, no-checkpoint execution time $T$, $E$ is the number of dependencies between layers, i.e. number of activations ($E=\Theta(L)$ in linear case, $E = O(L^2)$ in general) and $w$ is a treewidth of the computational graph.}
    \label{tab:remat}
\end{table*}

\paragraph{Support in popular open-source frameworks.}

Machine learning framework PyTorch~\footnote{\url{https://pytorch.org}} provides two options for checkpointing: the user explicitly defines which activations to store or uses a periodic strategy based on~\cite{Chen2016}.
Checkmate\footnote{\url{https://github.com/parasj/checkmate}}, written in TensorFlow\footnote{\url{https://www.tensorflow.org}}, accepts user-defined models expressed via the high-level Keras interface. Framework Rotor\footnote{\url{https://gitlab.inria.fr/hiepacs/rotor}} allows the algorithm from~\cite{herrmann2019optimal} to be used with any PyTorch DNN implemented with the {\tt nn.Sequential} container. 

\subsection{Offloading of Activations}

{\it Offloading} (also called {\it Memory Swapping}) is a technique that saves GPU memory by offloading activations to CPU memory during forward pass and prefetching them back into GPU memory for the corresponding backward computation.

Due to the limited bandwidth of the PCI bus between the CPU and the GPU, the choice of which activations to transfer and when to do it must be optimized. Authors of vDNN~\cite{Rhu2016} followed a heuristic effective for CNNs by offloading only the inputs of convolutional layers, though it does not generalize well to general DNNs. 
\cite{Le2018} considered the activations life-cycle to choose the candidates for offloading and used graph search methods 
to identify the instants when to insert offload/prefetch operations. AutoSwap~\cite{Zhang2019} decides which activations to offload by assigning each variable a priority score.
SwapAdviser \cite{Huang2020} used a Genetic Algorithm (GA) to find the best schedule (execution order of the modules) and memory allocation; it relied on Swap Planner to decide which tensors to offload (based on their life cycle) and when to perform offload/prefetch (as soon as possible).  
Authors in~\cite{Beaumont2020offloading} made a thorough theoretical analysis of the problem.
They proposed optimal solutions and extended them in~\cite{beaumont2021efficient} 
to jointly optimize activation offloading and rematerialization.

\begin{table*}[t]
    \centering
    \resizebox{\linewidth}{!}{
    \begin{tabular}{lp{0.25\textwidth}rp{0.35\textwidth}}
        \toprule
        Paper   & What to offload & Scope  & Features\\
        \midrule
vDNN \cite{Rhu2016} & all/conv & seq, CNN  & choice between memory/performance efficient kernels for conv \\
TFLMS \cite{Le2018} & tensors with longer lifetime & any & rewriting graph with swap in/out; treesearch with bounds to schedule transfers\\
AutoSwap \cite{Zhang2019} & tensors with highest priority scores & any & uses bayesian optmization to mix priority scores; optimizes memory allocation \\
SwapAdviser \cite{Huang2020} & tensors with longer lifetime & any & memory allocation and scheduling operations done with Genetic Algorithm\\
\cite{Beaumont2020offloading} & chosen by greedy/dynprog & het. seq. & theoretical analysis, optimality proofs for relaxed models\\
rotor \cite{beaumont2021efficient} & chosen by dynprog & het. seq. & combination of offloading and rematerialization, optimal w/ assumptions \\
        \bottomrule
    \end{tabular}
    }
    \caption{Comparison of offloading strategies.}
    \label{tab:off}
\end{table*}

\paragraph{Support in popular open-source frameworks.}
Framework vDNN++\footnote{\url{https://github.com/shriramsb/vdnn-plus-plus}} implemented  a technically improved version of vDNN. 
TFLMS~\cite{Le2018} was initially released as a TensorFlow pull request but later got its own repository\footnote{\url{https://github.com/IBM/tensorflow-large-model-support}}. A branch of the Rotor framework\footnote{\url{https://gitlab.inria.fr/hiepacs/rotor/-/tree/offload-all-rl}} provides the implementation of the combined offloading and rematerialization algorithms from~\cite{beaumont2021efficient}.

\subsection{Offloading of Weights}

A lot of methods mentioned earlier are also suitable for offloading weights as they rely on universal techniques applicable to any tensors, for example, TFLMS, AutoSwap or SwapAdvisor. 
L2L~(layer-to-layer)~\cite{pudipeddi2020training} keeps a single layer in GPU memory, which results in a significant reduction in the memory cost of the network. 
Granular CPU offloading \cite{lin2021m6} extends this approach and
keeps a part of the network in GPU when there is enough memory. Their experiment shows that offloading only the first half of the network 
can significantly reduce the training time.
ZeRO-Offload~\cite{Ren2021} managed to reduce the high communication cost by introducing the one-step Delayed Parameter Update (DPU) method. In ZeRO-Offload, only the gradients are offloaded to the CPU to update the weights and asynchronous updates are used to limit synchronizations. 

\paragraph{Support in popular open-source frameworks.}
DeepSpeed~\cite{rasley2020deepspeed} implements the extremely aggressive memory management strategies proposed in ZeRO-Offload~\cite{Ren2021}, which allows a single GPU to train models with more than 10 billion parameters.

\section{Parallelism for Models that Don't Fit on a Single GPU}\label{sec:parallelisms}

\begin{table*}[t]
    \centering
    \resizebox{\linewidth}{!}{
    \begin{tabular}{lrp{0.25\textwidth}p{0.25\textwidth}}
        \toprule
        Paper  & Parallelism & Pipeline Feature & Partition Optimization\\
        \midrule
        GPipe   \cite{gpipe}       & DP, PP & First introduced pipelining & - \\
        Megatron-LM \cite{megatron-pipeline}   & TP, DP, PP & 1F1B, Interleaved Pipeline & Heuristic \\
        PipeDream   \cite{pipedream}   & DP, PP & Async Update & DynProg for DP, PP\\
        PipeDream-2BW \cite{pipedream-2bw} & DP, PP  & Async Double-Buffered Update & DynProg for DP, PP, Checkpointing \\
        DAPPLE    \cite{dapple}     & DP, PP & 1F1B & DynProg for DP, PP \\
        PipeMare     \cite{pipemare}    & PP & Async Update, LR Rescheduling, Weight Discrepancy Correction  & Splitting weights evenly \newline between model partitions\\
        Piper   \cite{piper}      & TP, DP, PP & Async Update & DynProg for TP, DP, PP, Checkpointing \\
        HetPipe   \cite{hetpipe}     & DP, PP & Parameter Server & LinProg for PP \\
        Pipe-torch   \cite{pipe-torch}     & DP, PP & Async Update & DynProg for DP, PP, GPU allocation\\
        Varuna    \cite{Varuna}  & DP, PP & Opportunistic Backward Scheduling & Heuristic PP partition,\newline Bruteforce for DP, PP depth \\
        Gems    \cite{gems}  & DP, PP & Bidirectional Pipeline & - \\
        Chimera    \cite{Chimera}    & DP,PP & 1F1B, Bidirectional Pipeline & Greedy mini-batch size,\newline Bruteforce for DP, PP depth\\
        \bottomrule
    \end{tabular}
    }
    \caption{Comparison of model parallelism strategies.}
    \label{tab:pipelines}
\end{table*}


When using Model Parallelism \cite{Dean2012}, different layers of a network are allocated onto different resources, so that the storage of DNN weights and activations is shared between the resources. In Model Parallelism (MP), only activations have to be communicated and transfers only take place between successive layers assigned to different processors. Comparison of papers mentioned in this section is presented in \autoref{tab:pipelines}.

The execution within Model Parallelism can be accelerated if several mini-batches are pipelined~\cite{gpipe}, and thus several training iterations are active at the same time. 
Once forward and backward phases have been computed on all these mini-batches, the weights are then updated. 
This approach is fairly simple to implement but 
it leaves computational resources largely idle. 
The PipeDream approach proposed in~\cite{pipedream} improves this training process, by only enforcing that the forward and backward tasks use the same model weights for a given mini-batch. 
Such a weakened constraint on the training process allows PipeDream to achieve a much better utilization of the processing resources, but the asynchronous updates affect badly the overall convergence of the training in some cases~\cite{Chimera}.

It has been shown that performing the updates less regularly
\cite{pipedream-2bw} 
helps limiting weight staleness as well. 
Alternatively, PipeMare~\cite{pipemare} proposes to adapt the learning rate and the model weights for backward depending on the pipeline stage.
The last method achieves the same convergence rate as GPipe, while having the same resource utilization as PipeDream without storing multiple copies of the weights.
Another important issue related to PipeDream is the need to keep many copies of the model parameters, which can potentially cancel the benefit of using Model Parallelism. To address this issue, the methods to limit weight staleness can be used: in~\cite{pipedream-2bw} the updates are done so that it is possible to keep only two versions of the weights (Double-Buffering).

Modeling the storage cost induced by activations in pipeline approaches is a difficult task~\cite{beaumont2021pipelined}. Some pipelines~(DAPPLE \cite{dapple}, Chimera \cite{Chimera}) use the One-Forward-One-Backward scheduling (1F1B) to reduce memory consumption related to activations. It is a synchronous weight update technique that schedules backward passes of each micro-batch as early as possible to release the memory occupied by activations. 
Gems \cite{gems} and Chimera \cite{Chimera} implement bidirectional pipelines, where each GPU is used for two pipeline stages ($i$ and $P-i$, $P$ is the number of stages).
The design of Gems is mostly concerned with activations memory: the forward pass of the next micro-batch starts after the first backward stage of the previous micro-batch is computed and activations memory is released. Chimera rather focuses on reducing the computational bubble by starting the forward passes of each pipeline direction simultaneously. A resembling approach was taken in~\cite{megatron-pipeline}, where each GPU is assigned more than one pipeline stages (referred to as the Interleaved Pipeline).


Several papers specifically target challenging topologies. 
To solve the problem in the case of high communication costs and heterogeneous networking capabilities, the authors of Pipe-torch~\cite{pipe-torch} propose an updated dynamic programming strategy which assumes no overlap between computations and communications. 
HetPipe~\cite{hetpipe} addresses the additional problem of heterogeneous GPUs by grouping them into virtual workers and running pipeline parallelism within each virtual worker, while relying on data parallelism between workers. 
Varuna~\cite{Varuna} focuses on "spot" (low-priority) VMs and builds a schedule that is robust to network jitter, by performing a pipelining technique that resembles 1F1B: activation recomputations and respective backward passes are scheduled opportunistically. 


\section{Optimizers for Cross-Device Model Training}
\label{sec:optimizers}


\subsection{Zero Redundancy Optimizer}

The  authors of~\cite{Rajbhandari2020} propose ZeRO (Zero Redundancy Optimizer) as an implementation of data-parallelism with reduced memory footprint. 
The algorithm has three versions depending on what tensors are partitioned across devices: Stage 1 (optimizer states), Stage 2 (optimizer states and gradients), and Stage 3 (optimizer states, gradients and model parameters). ZeRO works in a mixed precision regime to reduce the amount of data transferred between devices. 
Still, Stages 2 and 3 introduce a communication overhead. 
In~\cite{Ren2021} authors propose to unite ZeRO and CPU-side computation of parameter updates within ZeRO-Offload: gradients are transferred to CPU where copies of parameters are stored; the update is applied to the copies and the updated weights are transferred back to GPU. 
\cite{Rajbhandari2021} further elaborates ZeRO-Offload with utilisation of NVMe memory, offload of activations, better computation-communication overlap and other improvements. 
\paragraph{Support in popular open-source frameworks.} An open source implementation of all ZeRO-* algorithms is available in the DeepSpeed\footnote{\url{https://github.com/microsoft/DeepSpeed}} framework.

\subsection{Low-Precision Optimizers}
To further reduce memory footprint, low-precision optimizers can be used.
These methods use low precision formats to represent the optimizers states and auxiliary vectors of states.
Also, error compensation techniques are used to preserve the approximation accuracy of the tracking statistics.
\cite{Dettmers21} proposes a method to store statistics of Adam optimizer in 8-bit while the overall performance remains the same as when the 32-bit format is used.
The key technique to achieve such a result is blockwise dynamic quantization that efficiently handles both large and small magnitude elements.
More aggressive precision reduction is presented in~\cite{sun2020ultra}, where special routines to deal with 4-bit representation are developed.
In particular, the adaptive Gradient Scaling (GradScale) method aims to mitigate the issue with insufficient range and resolution.


\subsection{Acceleration of Convergence}

Another way to accelerate training of large deep learning models is to reduce communication time between nodes and/or number of required epochs to converge at the appropriate local minimum.
\paragraph{Communication costs reduction.}
Different approaches have been proposed to compress gradients before transferring them between computational nodes.
In particular, three classes of such methods are typically discussed: sparsification, quantization and low-rank methods.
Sparsification methods only transfer some subset of complete gradient elements and update the corresponding elements in the parameter vector.
This approximation significantly reduces the communication costs~\cite{aji2017sparse,alistarh2019convergence} while the trained model performance is preserved.
Another approach is based on quantization of transferred gradients, which consists of transferring only a limited number of bits, reconstructing the entire gradient vector from them, and updating all elements of the parameter vector. 
This approach demonstrates promising results for some neural networks architectures and experimental settings~\cite{alistarh2017qsgd}.
In particular, recent results in sending only the signs of stochastic gradient elements~\cite{stich2018sparsified} have been extended to more complicated Adam optimizer~\cite{tang20211bit}, where the non-linear effect of the optimizer states requires additional investigation of error compensation strategies.  
Another approach for communication costs reduction is the low-rank approach, in which a low-rank approximation of the gradient is constructed, transferred and used to recover the gradient in full format before updating the parameter vector.
The low-rank approximation is constructed with the block power method~\cite{vogels2019powersgd} or with the alternating minimization strategy~\cite{cho2019gradzip}.
The main difficulty here is to balance the gains from the reduction of communication costs with the additional costs induced by the construction of low-rank approximations.
A comprehensive analysis and numerical comparison of many methods for communication overhead reduction from the aforementioned classes are presented in review~\cite{xu2021grace}.

\paragraph{Large batch training}
Another approach to speed up the convergence of the optimizer is to use a large number of samples per batch.
This training setting leads to a reduction of the number of iterations in every epoch and a better utilization of GPU.
In~\cite{goyal2017accurate} authors propose to use a linear scaling rule to update the learning rate along with the batch size. 
This setting stabilizes the optimization process and converges to the same final performance of the model.
Note also that large batch training significantly reduces the variance in the stochastic gradient estimate.
However, study~\cite{keskar2016large} 
observes that this feature of the training reduces the generalization ability of the trained model if other hyper-parameters remain the same.
Therefore, other alternatives to the linear scaling rule have been considered in further works.

In particular, Layer-wise Adaptive Rate Scaling is combined with SGD~\cite{you2017scaling} (LARS scheme) and Adam optimizers~\cite{you2019large} (LAMB scheme) to adjust the learning rate in every layer separately.
These strategies are based on the observation of a significant difference in the magnitude of parameters and gradients in the different layers, which is natural for deep neural networks.
Note that the memory saving techniques considered in Sections~\ref{sec:one_gpu} typically allow the batch size to be increased with little overhead, even when exceeding the GPU memory capabilities.



\section{Conclusion and Further Research}
In the survey we have discussed methods that help to train larger models on a single GPU and do more efficient training on multiple GPUs. Such methods optimize both the training of known high-quality large models (e.g., GPT, CLIP, DALLE) from scratch and the fine-tuning of pre-trained models for specific and personalized tasks.

Several main factors influence the training of large  DNNs:
(i) memory required to store model parameters, activations, optimizer states
(ii) time spent on data exchange (communication) between computing nodes and its impact on the computing time on a separate computing node, (iii) efficiency 
parallel computing 
(percentage of time when GPUs are not idle),
(iv) the number of floating-point operations required to calculate the forward 
and backward 
passes 
for the given model architecture, dataset, and target functionality, 
(v) the number of iterations required to achieve the specified accuracy. 
The strategies discussed in Sections~\ref{sec:one_gpu},\ref{sec:parallelisms}, and \ref{sec:optimizers} of this survey are applied to reduce the influence of these factors.



Current research in rematerilization (Table~\ref{tab:remat}) focus on finding the optimal checkpointing strategy if we have a homogeneous or heterogeneous sequential model. However, modern neural networks have a more complex structure - for example, due to a large number of residual connections. Currently, optimal rematerialization strategies for each architecture and input data size are heuristically searched. Further research can find theoretically optimal solutions for more general types of architectures.
Rematerialization methods demonstrated their benefits in reducing memory on one GPU. Despite that, it is important to combine it with other methods to achieve significant decrease in memory consumption. For example, in \cite{beaumont2021efficient} considering the optimal combinations of offloading and rematerialization (Section~\ref{sec:one_gpu}) further pushed the performance of both methods. Considering other combinations, e.g. checkpointing and pipelining (Section~\ref{sec:parallelisms}), can be a promising further development of both methods.


In optimization methods (Section~\ref{sec:optimizers}) there are three main research directions to adapt them for large model training: low-precision storage of states and gradients, batch size increasing with learning rate scheduling, compression of transferred gradients. They demonstrate promising results to train particular models 
but, at the same time, they are quite far from the complete technology. For example, the low precision approach requires extensive hardware support of the operations with numbers in a low-precision format, and the compression scheme for gradient transmission can be efficient only after the careful setting of the broadcasting environment. Thus, these 
approaches 
require additional research to make them robust and widespread.

Among the promising directions, we should mention computations with reduced precision, approximate methods~\cite{novikov2022few}, randomized computations~\cite{bershatsky2022memory}, and structured NN layers~\cite{hrinchuk-etal-2020-tensorized}, including those based on tensor factorizations.
We should highlight the importance for these approximate methods to be additive in the sense that they can be combined and still provide sufficient enough performance with reasonable quality degradation.

Finally, it is worth emphasizing the importance of developing new promising computing architectures that can speed up elementary machine learning operations (for example, matrix-vector multiplication) and low-level optimization techniques.



\section*{Acknowledgments}
The work was supported by the Analytical center under the RF Government (subsidy agreement 000000D730321P5Q0002, Grant No. 70-2021-00145 02.11.2021).\\
This work was supported by the Inria Molière International Associated Team between Skoltech and Inria. 
\newpage
\clearpage
\bibliographystyle{apalike}
\bibliography{
    biblio/intro.bib,
    biblio/checkoffload.bib,
    biblio/pipelines.bib,
    biblio/optimizers.bib,
    biblio/others.bib
}

\end{document}